# Clustering-Based Evolutionary Federated Multiobjective Optimization and Learning

Chengui Xiao, Songbai Liu*

College of Computer and Software Engineering, Shenzhen University, Shenzhen, China
2310273039@email.szu.edu.com, songbai@szu.edu.cn

**Abstract.** Federated learning enables decentralized model training while preserving data privacy, yet it faces challenges in balancing communication efficiency, model performance, and privacy protection. To address these trade-offs, we formulate FL as a federated multi-objective optimization problem and propose FedMOEAC, a clustering-based evolutionary algorithm that efficiently navigates the Pareto-optimal solution space. Our approach integrates quantization, weight sparsification, and differential privacy to reduce communication overhead while ensuring model robustness and privacy. The clustering mechanism enhances population diversity, preventing premature convergence and improving optimization efficiency. Experimental results on MNIST and CIFAR-10 demonstrate that FedMOEAC achieves 98.2% accuracy, reduces communication overhead by 45%, and maintains a privacy budget below 1.0, outperforming NSGA-II in convergence speed by 33%. This work provides a scalable and efficient FL framework, ensuring an optimal balance between accuracy, communication efficiency, and privacy in resource-constrained environments.

## 1 Introduction

Federated learning (FL) has emerged as a promising distributed machine learning paradigm, enabling multiple participants to collaboratively train a shared model while keeping their data localized. Unlike traditional centralized learning methods, which require direct data sharing, FL enhances data privacy by decentralizing computation across edge devices or clients [1]. Each client trains a local model and transmits only model updates to a central server, where global model aggregation occurs before redistributing updated parameters back to clients [2]. This decentralized approach has demonstrated significant potential in privacy-sensitive applications, such as healthcare [3] and finance [4], as well as in resource-constrained environments like mobile devices and the Internet of Things [5].

Despite its advantages, FL faces critical challenges, particularly in terms of communication overhead and privacy protection. Traditional federated optimization techniques often treat model training as a single-objective problem, primarily focusing on minimizing model loss [6]. However, real-world FL deployments involve multiple conflicting objectives: minimizing model loss to improve accuracy, reducing communication overhead to enhance efficiency, and strengthening privacy protection to mitigate data leakage risks [7]. Balancing these objectives is crucial for achieving an optimal trade-off in federated learning systems. Conventional FL optimization methods typically prioritize model performance while overlooking the trade-offs associated with communication costs and privacy risks [9]. The iterative nature of FL requires frequent model updates, leading to substantial data transmission between clients and the server [8]. In large-scale distributed environments, this results in high bandwidth consumption and prolonged communication delays, which hinder real-time model convergence. Moreover, despite keeping data localized, FL remains vulnerable to privacy threats such as inference attacks, where adversaries attempt to reconstruct original data by analyzing shared model gradients or parameters [10]. These challenges necessitate a shift from single-objective to a more comprehensive multiobjective optimization approach.

Given the inherent conflicts among model performance, communication efficiency, and privacy protection, FL should be viewed as a multiobjective optimization problem (MOP). Instead of optimizing a single metric at the expense of others, multiobjective optimization seeks to find a set of Pareto-optimal solutions that balance competing objectives [11]. By formulating FL as an MOP, we aim to derive optimal trade-offs that enhance learning efficiency while ensuring robust privacy safeguards. However, solving an MOP in FL is challenging due to the decentralized nature of training, heterogeneous data distributions across clients [12], and the dynamic constraints imposed by communication and computational limitations. Addressing these complexities requires an optimization method that can efficiently navigate the trade-offs among competing objectives while maintaining solution diversity.

Evolutionary algorithms (EAs) have proven highly effective in solving MOPs due to their ability to explore diverse solution spaces and identify optimal trade-offs among conflicting objectives [13]. Unlike traditional gradient-based methods, which often struggle with local minima and require extensive parameter tuning, EAs employ population-based search strategies to evolve multiple solutions in parallel [14]. This allows them to approximate the Pareto front efficiently, ensuring a well-balanced compromise between competing objectives. In the context of FL, where model performance, communication efficiency, and privacy protection must be optimized simultaneously, EAs provide a natural and flexible approach for achieving these goals without being constrained by strict mathematical formulations [15].

To address the federated MOP challenge, we propose a clustering-based EA that enhances the effectiveness of the optimization process. One of its key advantages is diversity maintenance [16], as clustering helps preserve a broad set of Pareto-optimal solutions by grouping similar solutions together, preventing premature convergence to suboptimal points [17]. Additionally, the exploration-exploitation trade-off is improved by adapting search strategies within clusters, allowing for thorough exploration of the solution space while simultaneously refining high-quality solutions [18]. This ensures that the algorithm can effectively adapt to different FL environments, where client models and data distributions vary significantly. Furthermore, the scalability of the clustering-based approach makes it particularly effective for large-scale FL systems. By grouping clients with similar optimization tendencies, the algorithm reduces redundant computations and accelerates convergence without sacrificing solution quality. This hierarchical structuring also mitigates the computational burden on individual clients, making the approach more practical for real-world FL deployments. Through this method, we achieve an efficient and adaptive federated learning framework that balances accuracy, communication efficiency, and privacy protection, providing a robust alternative to traditional single-objective optimization strategies.

To address the federated MOP, our work makes the following key contributions:

- we redefine FL as a multiobjective optimization problem, considering the simultaneous minimization of model loss, communication overhead, and privacy leakage.
- Our approach comprehensively incorporates the three key FL factors, providing a holistic optimization perspective.
- We introduce a clustering-based EA, termed FedMOEAC, to efficiently find optimal trade-off solutions within the FL framework.

The remainder of this paper is structured as follows. The next section introduces the optimization objectives in FL and discusses the motivation behind our study. We then detail the integration of the proposed algorithm, followed by experimental validation. Finally, we summarize our findings and outline future research directions.

## 2 Related Work and Motivation

FL presents a unique paradigm that balances decentralized model training with privacy preservation. However, its deployment is constrained by multiple conflicting objectives, particularly communication efficiency, model performance, and privacy protection. Traditional FL optimization methods largely prioritize model accuracy while neglecting the inherent trade-offs among these factors. Consequently, recent advancements in FL have embraced a multi-objective optimization perspective to derive optimal trade-off solutions that enhance efficiency, robustness, and fairness.

### 2.1 Communication Overhead in Federated Learning

Frequent communication between clients and the central server is a fundamental bottleneck in FL, leading to excessive bandwidth consumption and prolonged model convergence. Mitigating this communication overhead is essential to making FL viable for large-scale deployment. Strategies for addressing this issue typically include client selection, model compression, and reducing update frequency.

Client selection mechanisms optimize which participants contribute to model updates, thereby improving efficiency without compromising model quality. For instance, strategies that prioritize high-impact clients based on their gradient variance or model divergence accelerate FL convergence [19]. Meanwhile, reducing update frequency minimizes unnecessary transmissions by leveraging techniques such as adaptive dropout and knowledge distillation [20]. These approaches allow clients to update the global model less frequently while retaining knowledge transfer effectiveness. Additionally, model compression methods, including weight sparsification [21] and quantization [22], reduce the size of transmitted updates, thereby lowering communication costs. However, optimizing communication efficiency in isolation can lead to performance degradation or increased privacy risks. For example, aggressive compression may im-pair model accuracy, while infrequent updates might prolong training. Solving a federated MOP enables the simultaneous optimization of communication efficiency, model accuracy, and privacy protection, ensuring a balanced trade-off.

### 2.2 Privacy Protection through Differential Privacy

Despite its decentralized nature, FL remains vulnerable to privacy attacks, including inference attacks [23] and gradient leakage [24]. Differential privacy (DP) is a widely used mechanism to mitigate these threats by introducing carefully calibrated noise into shared model updates [25]. DP ensures that adversaries cannot infer individual data points by analyzing aggregated updates [26]. In federated optimization, DP is often implemented using the Gaussian mechanism [27], where noise is added to gradients before aggregation. However, an excessive noise scale can degrade model accuracy, creating a trade-off between privacy and performance. Differentially private stochastic gradient descent addresses this by applying gradient clipping before noise injection, ensuring that individual client updates remain within a bounded range. The privacy budget, which controls the level of protection, must be carefully managed to maintain a balance between learning effectiveness and privacy preservation.

From a federated MOP perspective, DP must be integrated into the optimization process alongside model performance and communication efficiency. Traditional single-objective approaches may overemphasize privacy at the cost of accuracy, while federated MOP ensures that privacy-enhanced solutions maintain acceptable levels of model utility and communication feasibility.

### 2.3 Multiobjective Optimization in Federated Learning

Given the inherent conflicts among accuracy, communication efficiency, and privacy, FL is naturally suited for a multiobjective optimization framework. Recent studies have explored various combinations of these objectives to develop more adaptive FL strategies. For instance, constrained multi-objective FL frameworks integrate privacy leakage, training cost, and utility loss within a single optimization process, leveraging EAs such as NSGA-II to identify Pareto-optimal solutions [28]. Other approaches have incorporated fairness constraints [29], ensuring that client contributions are equitably weighted in model updates. Additionally, research has demonstrated the benefits of robust multiobjective optimization [30] by incorporating adversarial resilience, mitigating vulnerabilities while maintaining efficiency.

EAs play a crucial role in solving federated multiobjective optimization challenges. Unlike traditional gradient-based optimization, which struggles with local optima and requires extensive parameter tuning, EAs explore diverse solution spaces, enabling effective trade-offs. Specifically, clustering-based EAs have proven particularly effective, as they partition solutions into groups based on cosine similarity, maintaining diversity and preventing premature convergence. Unlike Euclidean-based approaches, cosine similarity better captures population distribution in high-dimensional FL settings, leading to more robust optimization outcomes. To quantify the effectiveness of multiobjective FL approaches, performance is typically evaluated using hypervolume (HV), which measures the space covered by Pareto-optimal solutions [31]. This metric ensures that solutions maintain diversity while optimizing multiple conflicting criteria.

By framing FL as a federated MOP, we redefine the training paradigm to account for accuracy, communication efficiency, and privacy protection. Our approach leverages clustering-based evolutionary algorithms to efficiently navigate these trade-offs, ensuring that FL systems achieve optimal performance without compromising security or efficiency.

## 3   The Formulation of Federated MOP

In this work, the federated MOP is formulated to simultaneously optimize three key objectives in FL: global model accuracy, communication efficiency, and privacy protection. These objectives inherently conflict, necessitating a trade-off that balances model performance, resource constraints, and privacy guarantees. The formulation of the federated MOP is defined as follows:

$$\text{Minimize} \begin{cases} \text{Global error: } f^{ge}(x) \\ \text{Communication overhead: } f^{co}(x), \text{where } x = (\xi, q, \sigma) \\ \text{Privacy budget: } f^{pb}(x) \end{cases} \quad (1)$$

where $\xi$ is a predefined threshold that determines the level of weight sparsification, effectively reducing communication overhead between clients and the server. Specifically, the parameters of the fully connected layer in the $k$-th client are pruned using a sparsification operator, which sets weights with absolute values below $\xi$ to zero:

$$w_k^{\text{pruned}} = \begin{cases} w_k, & \text{if } |w_k| \geq \xi \\ 0, & \text{otherwise} \end{cases} \quad (2)$$

By eliminating negligible weights, the sparsification strategy reduces data transmission while preserving model accuracy, enhancing communication efficiency in FL.

In addition, $q$ represents the quantized bit-width in the quantization operator, which reduces the precision of the sparse model's weights while maintaining their representational capacity. This process further compresses the model by mapping continuous weights to their discrete counterparts, as follows:

$$\tilde{w}_k = \frac{\left\lfloor (2^q - 1) \cdot \frac{w_k^{\text{pruned}} - w_{\min}}{w_{\max} - w_{\min}} \right\rfloor}{(2^q - 1)} \cdot (w_{\max} - w_{\min}) + w_{\min} \quad (3)$$

where $w_{\min}$ and $w_{\max}$ are the minimum and maximum values of the weight range, respectively. The weights are first normalized to the range [0,1] and then mapped to discrete levels. This quantization process reduces storage requirements by converting 32-bit floating-point parameters into $q$-bit integers, effectively compressing the model while retaining essential information for performance.

Furthermore, $\sigma$ is the noise scale to control the amount of noise added in balancing privacy protection and model utility. Specifically, to enhance the model's privacy protection capabilities, Gaussian noise $G(0, \sigma^2 Z^2)$ is added to the local model before transmission to the server. This process is essential for differential privacy, ensuring that the inclusion or exclusion of any single data sample has minimal impact on the model's output, thereby protecting individual privacy. The perturbed parameter $\tilde{w}_k^{\text{noisy}}$ is:

$$\tilde{w}_k^{\text{noisy}} = average\left(\frac{\tilde{w}_k}{\max(1, \|\tilde{w}_k\|/Z)}\right) + G(0, \sigma^2 Z^2) \tag{4}$$

where $Z$ is the clipping threshold that bounds the gradient norms of the local model, preventing any single gradient from disproportionately influencing training. After clipping, Gaussian noise is added to further obscure individual contributions. The resulting noisy models are then transmitted to the server for aggregation, ensuring that global updates do not leak sensitive information, thereby strengthening privacy guarantees.

In this way, the server aggregates the compressed and privatized client models to optimize the global objective:

$$f^{ge}(x) = \sum_{k=1 \text{ to } K} L\left[D_k, M(\tilde{w}_k^{noisy}, x)\right] \tag{5}$$

where $K$ is the number of clients and $M(\tilde{w}_k^{noisy}, x)$ represents the $k$th local model after compression and adding noise, which is locally trained on its personalized data set $D_k$. Besides, $L[\circ]$ indicates the employed loss function to evaluate the error of the model, which is computed based on the uploaded weights $\tilde{w}_k^{noisy}$. Moreover, $x = (\xi, q, \sigma)$ denotes the parameters to be optimized. The total communication overhead is defined as:

$$f^{co}(x) = \frac{\sum_{k=1 \text{ to } K} \Gamma\left[M(\tilde{w}_k^{noisy}, x)\right]}{\sum_{k=1 \text{ to } K} \Gamma[M(w_k)]} \tag{6}$$

where $M(w_k)$ denotes the original model of the $k$th client without any processing and $\Gamma[\circ]$ measures the model size before and after compression. The privacy budget $f^{pb}(x)$, which quantifies privacy leakage, is given by:

$$f^{pb}(x) = \frac{\sqrt{2T \ln(1/\delta)}}{v\sigma} \tag{7}$$

where $v$ is the client sampling rate, $\delta$ is a small error margin, and $T$ is the number of global training rounds. A smaller value provides stronger privacy protection but may degrade model accuracy due to excessive noise. By optimizing $f^{ge}(x)$, $f^{co}(x)$, and $f^{pb}(x)$ jointly using a clustering-based multiobjective EA, we ensure that FL achieves a balanced trade-off between accuracy, communication efficiency, and privacy.

## 4   The Proposed Clustering-based Evolutionary Algorithm

This section presents the FedMOEAC framework, which optimizes conflicting objectives in federated learning using a clustering-based evolutionary approach. The algorithm begins by initializing a population of solutions that define neural network architectures. These solutions are distributed to clients, which train local models on their respective datasets. Based on aggregated global updates, the evolutionary process selects well-balanced solutions, ensuring improved model accuracy, communication efficiency, and privacy protection. The overall framework is illustrated in Fig. 1 and detailed in Algorithm 1.

The population $P$ is initialized using a Gaussian distribution defined by its mean and

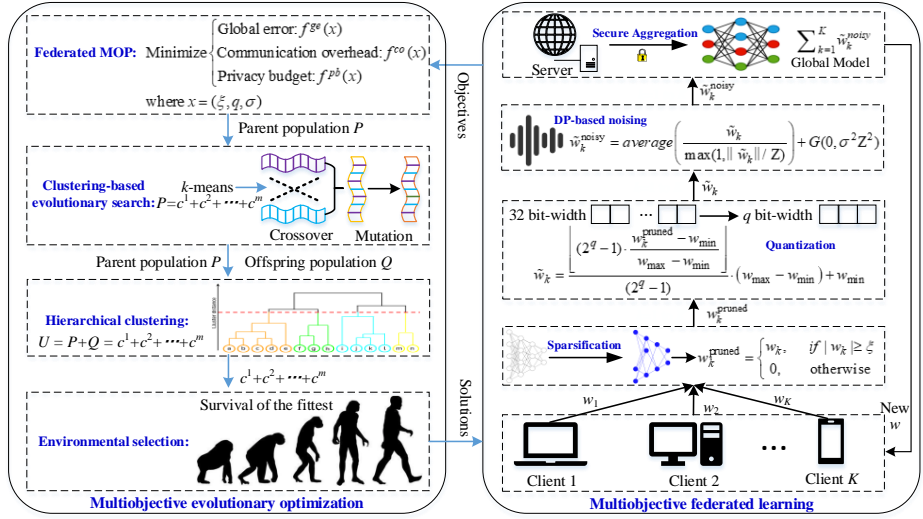

Fig. 1 Illustration of the general framework of the proposed FedMOEAC

---

**Algorithm 1**: General Framework of the proposed FedMOEAC
**Input:** population size $N$, maximum generations $G_{max}$, list of client datasets
**Output:** the final population $P$
 1:  initialize the population $P$ with $N$ random solutions;
 2:  global_model = initialize_global_model($P$);
 3:  **while** $G <= G_{max}$ **do**
 4:      divide $P$ into $m$ clusters by the $k$-means method;
 5:      crossover and mutate within clusters to get offspring population $Q$;
 6:      combine $P$ and $Q$ to get the union population $U$;
 7:      local_model = train_local_model(global_model, client.dataset);
 8:      global_model = aggregate_updates(local_models);
 9:      for each $x$ in $U$, $x$.fitness = evaluate(global_model, $f^{ge}$, $f^{co}$, $f^{pb}$);
10:      $(c^1, c^2, \ldots, c^N) \leftarrow$ Hierarchical_Clustering($U$, $N$);
11:      $P \leftarrow$ select the solution with the minimum fitness value from each cluster $c^i$;
12:      $G = G + 1$;
13: **end while**
14: **return** the final population $P$

---

variance. Each solution within the population is encoded by $x = (\xi, q, \sigma)$. Specifically, $q$ denotes the quantization bit-width, $\xi$ controls the sparsity of client models, and $\delta$ represents the DP-based noise level. After initialization, the global model uses these solutions to configure the network structure. FedMOEAC optimizes conflicting objectives (1) by employing a clustering-based evolutionary strategy. The initial population $P$ with $N$ solutions is partitioned into $m$ clusters by k-means based on cosine similarity in the objective space. This ensures diversity across clusters. Randomly selecting parents from each cluster to generate the offspring population $Q$ with $N$ new child solutions using evolutionary operators (e.g., crossover, mutation). Parent solutions are randomly selected within clusters to maintain diversity while preserving convergence properties.

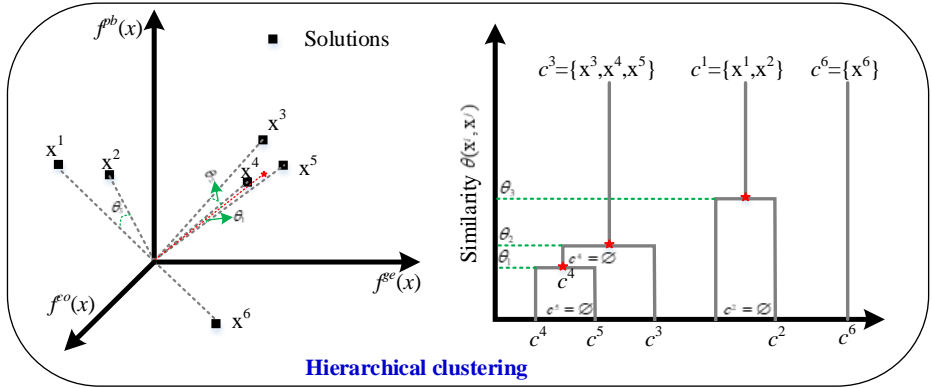

Fig. 2. A toy example to illustrate the process of the hierarchical clustering.

The parent population $P$ and the offspring population $Q$ are combined to form a unified population $U$. Each solution $x$ in $U$ is evaluated by the global error, communication overhead, and privacy budget. Environmental selection aims to minimize conflicting objectives by identifying a balanced solution. To achieve this, solutions in $U$ are grouped using a hierarchical clustering method, as a toy example shown in Fig. 2. The similarity between any two points is measured by their cosine similarity, with smaller values indicating closer directions. Initially, each individual is treated as a separate cluster. At each step, the two clusters with the highest cosine similarity are merged into a new cluster. This hierarchical clustering process continues until the number of clusters is reduced to $N$. The similarity between a solution $x$ and a cluster $c$ is defined as the cosine acute angle between $x$ and the cluster center of $c$. Besides, once two clusters are merged to get the new cluster $c^{new}$, the cluster center of $c^{new}$ is updated accordingly. After the clustering process, from each cluster, the solution with the minimum fitness value is selected for the new population $P$. This method ensures that the population maintains high diversity while also converging towards optimal solutions, effectively balancing the trade-offs between global error rate, communication overhead, and privacy budget. The process continues iterating until a predefined termination condition is satisfied.

## 5   Experimental Studies

### 5.1  Experiment Settings

This section describes the experimental settings used in our case study, which include the following aspects: 1) The neural network models employed and their configurations; 2) Federated learning parameter settings; 3) NSGA-II parameter settings; 4) FedMOEAC parameter settings. We selected two widely used neural network architectures: VGG and CNN. The VGG model was trained on CIFAR-10, while the CNN was used for MNIST. The VGG model utilized a mini-batch SGD optimizer with a learning rate of 0.01 and a batch size of 32. The CNN architecture consisted of two convolutional layers, two fully connected layers, and ReLU activation functions. The VGG model included multiple VGGBlocks, with fully connected layers for classification. To enhance stability and generalization, we incorporated residual connections and GroupNorm. These architectures serve as standard benchmark models in our experiments. For federated learning, we set the total number of clients $K = 10$, with a client participation rate of 0.4, meaning that 4 clients participated in each communication round. The minibatch size ($B$) and the number of local training epochs ($E$) were configured as MNIST (CNN): $B = 64$, $E = 2$, and CIFAR-10 (VGG): $B = 32$, $E = 5$.

Table 1. Performance comparison between FedMOEAC and NSGA-II across different datasets and models.

| Algorithms | Datasets | Models | $f^{to}(\%)$ | $f^{ge}(\%)$ | $f^{pb}(\%)$ |
|---|---|---|---|---|---|
| NSGA-II | MNIST | CNN | 54.19 | 1.95 | 0.4911 |
|  |  | VGG | 50.28 | 1.75 | 0.1865 |
|  | Cifar10 | CNN | 60.27 | 29.26 | 0.1228 |
|  |  | VGG | 46.37 | 28.35 | 0.0859 |
| Ours | MNIST | CNN | 31.29 | 1.65 | 0.1123 |
|  |  | VGG | 30.28 | 1.72 | 0.1121 |
|  | Cifar10 | CNN | 52.94 | 27.95 | 0.4203 |
|  |  | VGG | 45.69 | 27.75 | 0.0441 |

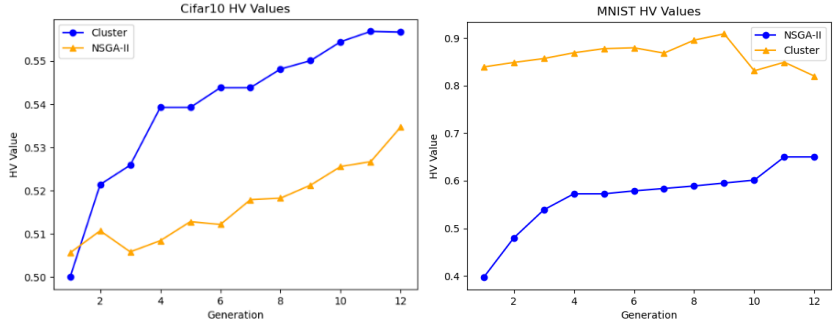

**Fig. 3.** HV value evolution of FedMOEAC and NSGA-II over generations.

Due to computational constraints, we set the population size to 10 and ran the optimization process for 12 generations. The crossover and mutation parameters for real-valued variables were empirically set as SBX crossover: probability 0.9, distribution index 2, and Polynomial mutation: probability 0.1, distribution index 20. For hypervolume (HV) calculation, we used a unit reference point with three components, corresponding to normalized error rate, communication overhead, and privacy budget.

The initial population was randomly generated from a Gaussian distribution. The quantization ratio was set to 1, 1/2, and 1/4 of the original bit precision, corresponding to 32-bit, 16-bit, and 8-bit quantization levels. To ensure strong privacy protection, we constrained the privacy budget within the range [0.1, 10]. The pruning rate was set with a mean of 0.3 and a standard deviation of 0.1, while the noise rate was set with a mean of 6 and a standard deviation of 2.5, ensuring a balanced trade-off between model accuracy and privacy preservation.

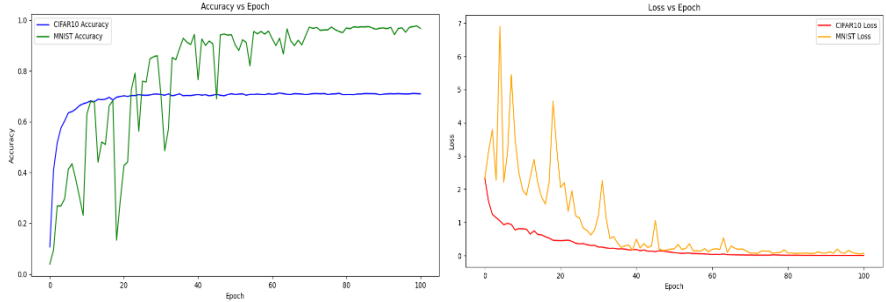

**Fig. 4.** The accuracy and loss of solution.

### 5.2 Experiments on FedMOEAC

The experimental results strongly demonstrate the superiority of FedMOEAC over NSGA-II in federated learning optimization. Table 1 provides a detailed comparison across different datasets and models, highlighting the consistent performance gains achieved by FedMOEAC. On the MNIST dataset with a CNN model, FedMOEAC reduces communication costs from 54.19% to 31.29%, lowers the privacy budget from 0.4911 to 0.1123, and improves the error rate from 1.95% to 1.65%. These improvements are not isolated cases but are consistently observed across all datasets and models, reinforcing the robustness and effectiveness of our approach in optimizing model performance, communication efficiency, and privacy protection.

As shown in Fig. 3, the HV values of the population steadily increase across generations, indicating strong convergence. FedMOEAC consistently achieves higher HV values than NSGA-II, reflecting its superior ability to maintain solution diversity. This balance between convergence and diversity is essential in multi-objective optimization, ensuring a comprehensive exploration of the Pareto front and enabling a broad range of trade-off solutions. The superior HV performance of FedMOEAC confirms its effectiveness in navigating the complex landscape of federated learning optimization.

Fig. 5 provides further insights into the quality and distribution of solutions obtained by FedMOEAC. The final population solutions exhibit higher accuracy and lower loss values, suggesting better model generalization to unseen data. Additionally, the solutions are well-distributed across the Pareto front, demonstrating the algorithm's capability to thoroughly explore the search space while avoiding premature convergence to suboptimal regions. The spatial distribution of solutions in Fig. 5 further highlights FedMOEAC's ability to explore a wide range of trade-offs among communication cost, error rate, and privacy budget. This diversity is particularly valuable in practical applications, where different stakeholders may prioritize different objectives. Notably, FedMOEAC effectively prevents solution clustering in suboptimal regions, ensuring that the final set of solutions fully covers the Pareto front.

The consistent and substantial improvements across all metrics, along with FedMOEAC's superior convergence and solution diversity, underscore its practical applicability and theoretical soundness in addressing the challenges of modern federated learning scenarios.

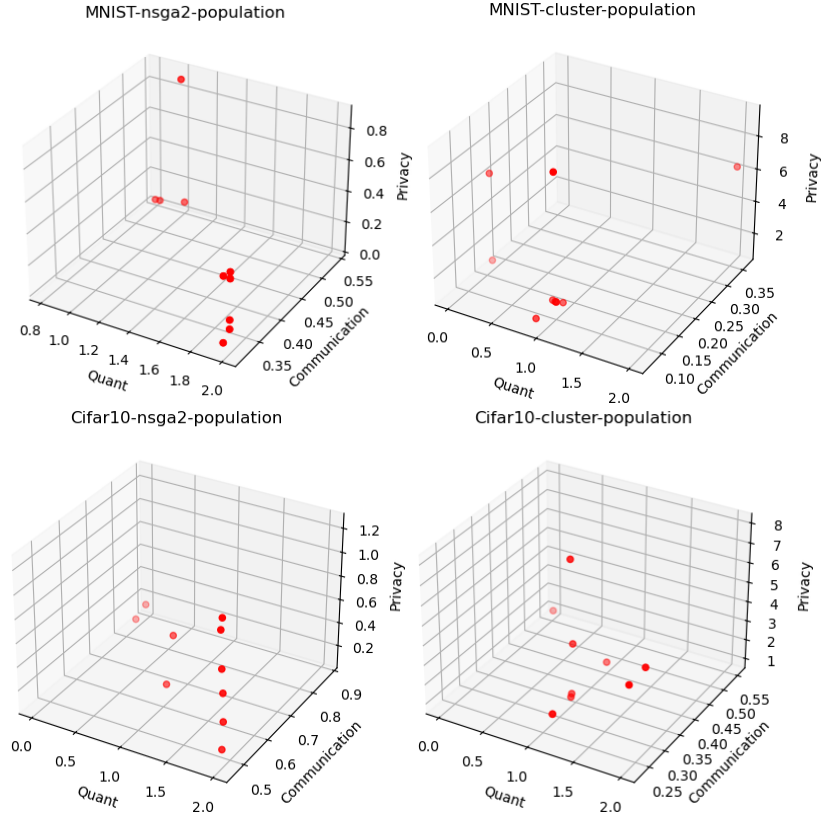

**Fig. 5.** Final population distribution of FedMOEAC on the Pareto front.

## 6   Conclusion

This paper presents FedMOEAC, a clustering-based evolutionary algorithm for federated multi-objective optimization, addressing the fundamental trade-offs in communication efficiency, model accuracy, and privacy protection. By incorporating quantization, weight sparsification, and differential privacy, the proposed framework significantly reduces communication costs while preserving model performance and privacy guarantees. The clustering-based evolutionary approach enhances solution diversity, mitigating local optima and accelerating convergence. Experimental evaluations on MNIST and CIFAR-10 validate the superiority of FedMOEAC over NSGA-II, achieving higher accuracy, lower communication overhead, and improved privacy efficiency. These findings highlight the potential of evolutionary multi-objective optimization in FL, offering a scalable and effective solution for real-world decentralized learning scenarios. Future research will explore extending FedMOEAC to heterogeneous and dynamic FL environments, further improving adaptability and robustness.